# Comparative Evaluation of Deep Learning Models for Geographic Atrophy Segmentation in En Face OCT Images

**Utkarsh Doshi**[1]

**Adarsh Gadari**[1]

**Sharat Chandra Vupparaboina**[1]

**Joseph DeCicco**[1]

**Giulia Gregori**[1, 4]

**Nicola Valsecchi, MD**[1,2,3]

**Nasiq Hasan, MD**[1]

**Elham Sadeghi, MD**[1]

**José-Alain Sahel, MD**[1]

**Jay Chhablani, MD**[1]

**Kiran Kumar Vupparaboina, PhD**[1]

**Sandeep Chandra Bollepalli, PhD**[1]

From:

[1] Department of Ophthalmology, University of Pittsburgh school of Medicine, Pittsburgh

[2] Ophthalmology Unit, Dipartimento di Scienze Mediche e Chirurgiche, Alma Mater Studiorum University of Bologna, Bologna, Italy.

[3] IRCCS Azienda Ospedaliero-Universitaria di Bologna, Bologna, Italy

[4] Department of Experimental and Clinical Medicine, Eye Clinic, Polytechnic University of Marche, Ancona, Italy

# 1. Abstract

Background: Geographic atrophy (GA) is an advanced form of age-related macular degeneration (AMD). It causes irreversible vision loss due to progressive degeneration of the retinal pigment epithelium (RPE) and photoreceptors. Accurate segmentation of GA in en-face image is crucial for early diagnosis, monitoring disease progression, and developing personalized treatment strategies. However, existing deep learning approaches are typically evaluated using single architectures and single-device datasets, limiting insight into relative performance and cross-device robustness.

Methods: We trained and systematically compared five deep learning architectures for GA segmentation: Nested U-Net, ResU-Net, nnUnet, Swin-Unet and MedSegDiff model, together with a thresholding based classical baseline. The models were trained on a dataset of en-face OCT images derived from 269 OCT volumes from two different devices, following a three-step preprocessing technique involving normalization and inpainting to enhance the quality of the en-face images.

Results: On the fixed held-out validation set comprising OCT scans from both Cirrus® (Carl Zeiss Meditec, Inc., USA) and Spectralis® (Heidelberg Engineering, Germany) devices, nnUnet achieved the highest mean Dice similarity coefficient (DSC) of 0.8882 ± 0.0033. The other models yielded the following Dice scores: Nested U-Net (0.7932 ± 0.0423), ResU-Net (0.8453 ± 0.0074), Swin-Unet (0.6886 ± 0.0294), and Diffusion model (0.7959 ± 0.0242).

Conclusions: nnUnet framework, demonstrated most reliable performance for GA segmentation in en-face OCT under limited data conditions. It offers the best balance of accuracy, stability, and efficiency for GA quantification.

# 2. Introduction

Geographic Atrophy (GA) is the advanced form of Age-related macular degeneration (AMD), which remains the leading cause of irreversible vision loss in older adults[1]. GA is characterized by progressive degeneration of the outer retinal layers, including retinal pigment epithelium (RPE) and photoreceptors, which are essential for capturing and encoding visual information[2,3]. Optical Coherence Tomography (OCT) is a light-based imaging technique that provides high-resolution, three-dimensional visualization of retinal sublayers, enabling detailed assessment of atrophic changes across retinal layers.

Accurate quantification of GA lesion size and growth rate is of increasing clinical importance. GA progresses gradually over time, and lesion growth rate has emerged as a key endpoint in therapeutic trials. Reliable segmentation of GA is therefore essential not only for monitoring disease progression and predicting visual prognosis, but also for assessing treatment response in clinical trials. Manual measurement of GA from OCT en-face is time-consuming, subject to inter-grader variability, and impractical at the scale required for large cohort studies or routine clinical monitoring. Automated and reproducible segmentation methods that can accurately delineate GA boundaries and quantify lesion area across imaging sessions and devices are therefore a critical unmet need in the clinical management of AMD.

Degeneration of RPE allows increased light penetration into the choroid, manifesting as hypertransmission on the OCT B-Scans, commonly referred to as Increased Choroidal Transmission (ICT). Qualitative assessment of GA based on such features is prone to observer variability, and the manual segmentation of retinal layers to quantify GA in OCT volumes is labor-intensive, highlighting the need for automated, reliable, and reproducible segmentation techniques[4]. Most existing automated GA quantification methods use Fundus Autofluorescence (FAF) or OCT en-face images. However, hypoautofluorescent regions in FAF are not specific to GA and may arise from other retinal pathologies, which can complicate automated differentiation[5]. To address this limitation, several studies have proposed exploiting ICT observed in OCT B-Scans by generating en-face projections (Fig 1) of the choroid region beneath the RPE. In this study, we created en-face images using a choroid segmentation model[6] (Fig 2a) and compared the performance of different deep learning models on segmenting the GA region on an en-face image.

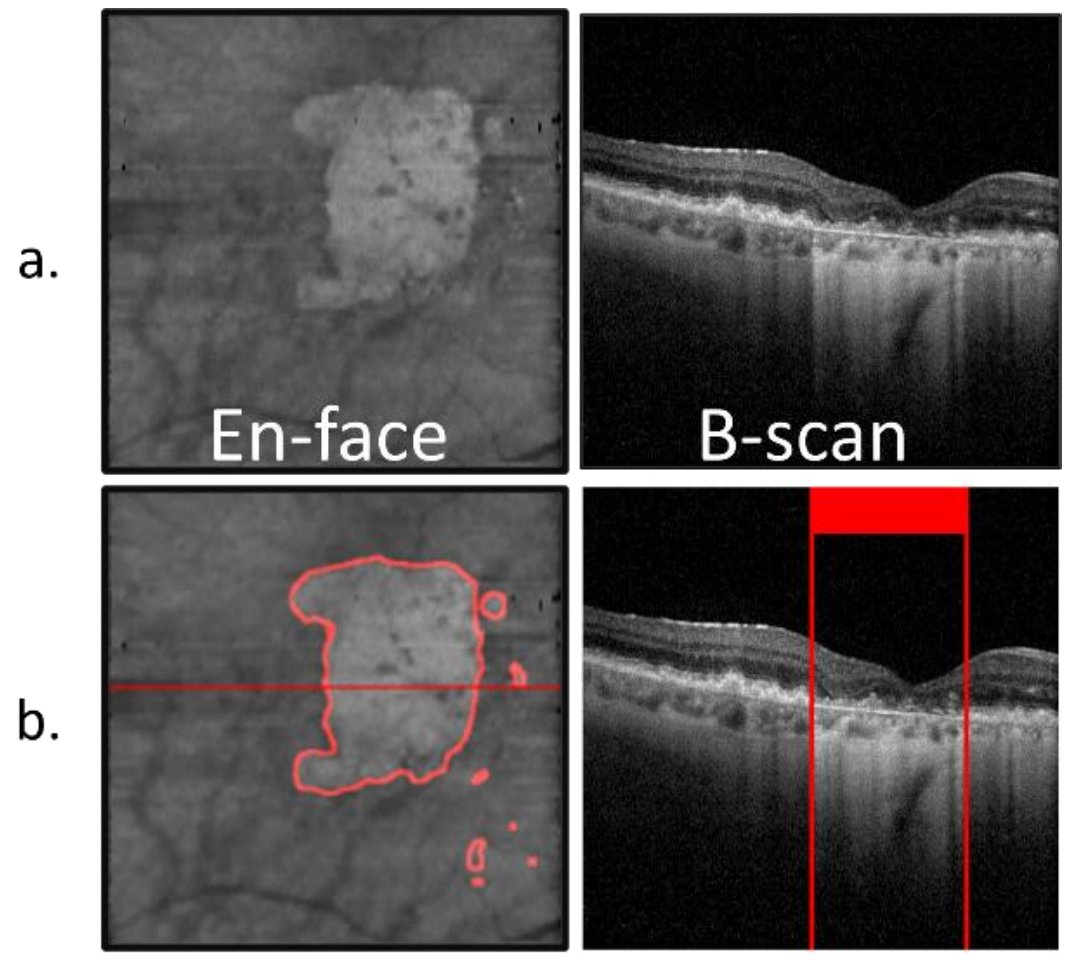

**Fig 1. (a)** En-face image generated by averaging intensity values from a 100-pixel slab located beneath the choroid top layer across all B-scans. (b) Geographic atrophy (GA) annotations overlaid on the en-face image, with a horizontal red line denoting the location of the corresponding B-scan.

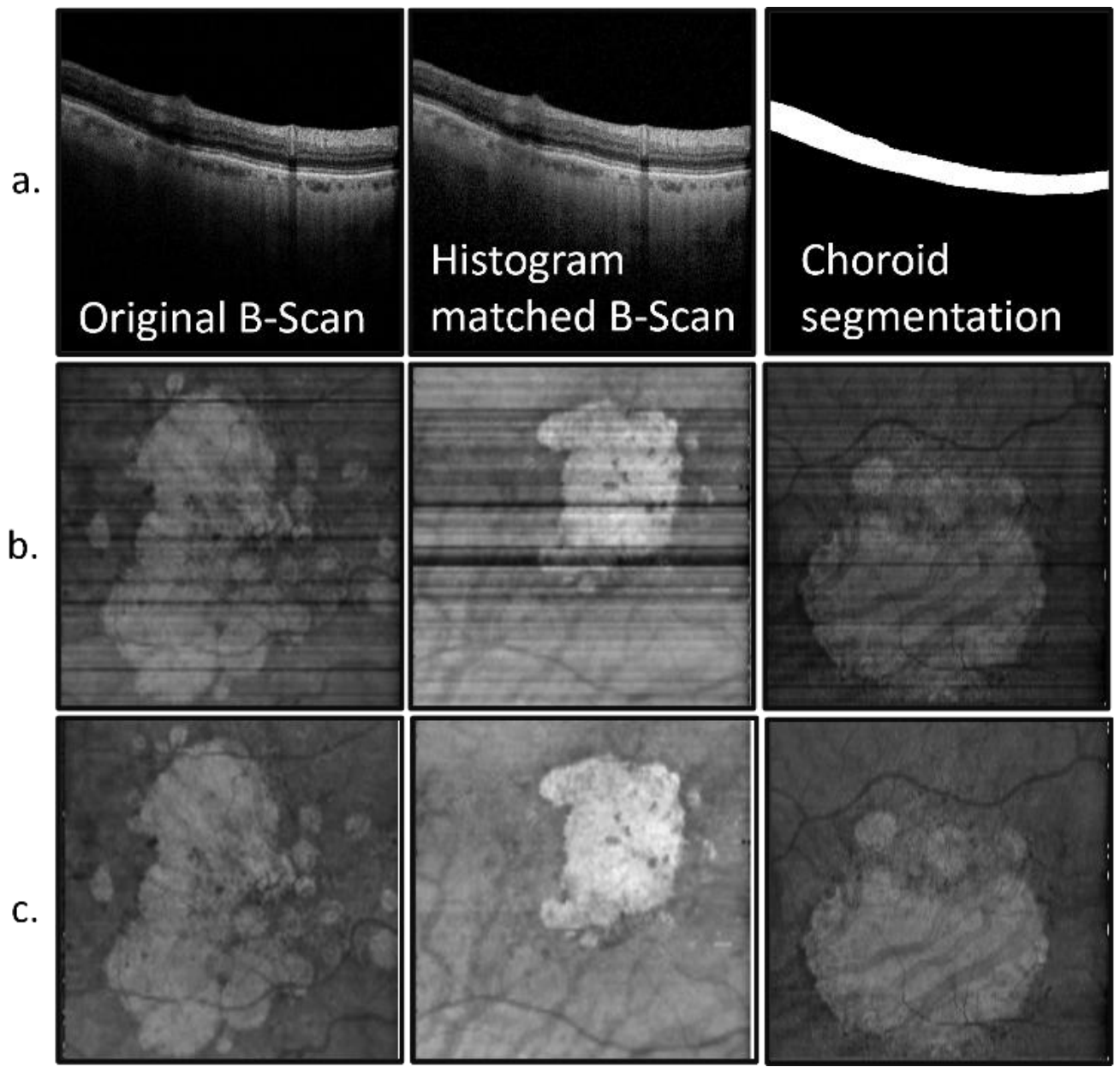

**Fig 2a** The histogram of each original B-scan is matched to that of a reference B-scan prior to choroid segmentation.

Deep learning, particularly convolutional neural networks (CNNs)[7], has emerged as a widely adopted approach for medical image segmentation[8]. U-Net[9] and its derivatives have shown strong performance across various segmentation tasks, including retinal images like OCT. More recently, alternative architectures such as transformer-based models[10], and diffusion-based segmentation models[11] have gained traction for medical image segmentation, each offering unique strengths in capturing fine structures. Despite these advances, systematic comparisons of such architectures for GA segmentation on OCT-derived en-face images remain limited, particularly under realistic data constraints.

In this study, we evaluate GA segmentation performance on OCT-derived en-face images using multiple deep learning architectures, including U-Net variants (nnUNet[12], ResU-Net[13], and Nested U-Net[14]), the transformer-based Swin U-Net[15], and the diffusion-based MedSegDiff[16]. A key challenge in GA segmentation arises from imaging artifacts like inter-slice contrast variability, B-scan misalignment, and motion-induced distortions, which are further exacerbated by differences in acquisition protocols across OCT devices. For example, cirrus (Zeiss) and Spectralis (Heidelberg) systems differ substantially in scan resolution and intensity characteristics, often limiting the generalizability of the automated methods to a single device or protocol.

To address these challenges, we introduce a standardized en-face image reconstruction pipeline incorporating choroid segmentation, histogram matching, and artifact inpainting, enabling fair and consistent model comparison. Using this framework, we systematically benchmark multiple deep learning architectures for GA segmentation under limited data conditions. We believe this work provides practical insights into architecture-level trade-offs and contributes toward standardized evaluation of GA segmentation methods using OCT derived en-face representations, supporting future development of clinically applicable AI tools for retinal diseases monitoring and progression analysis.

# 3. Related works

Early methods for segmenting GA primarily relied on classical image processing and threshold-based algorithms. For instance, region-growing techniques combined with morphological operations were applied to fundus autofluorescence (FAF) images, yielding moderate overlap with manual annotations despite variability in lesion contrast and imaging noise[17]. Similarly, summed-voxel projection (SVP) methods were used to generate projection images from spectral-domain OCT (SD-OCT) volumes, followed by thresholding and contour-based approaches to delineate GA regions[18]. Although computationally efficient, these traditional approaches were limited in robustness and struggled to capture the morphological complexity of GA lesions, especially in the presence of imaging artifacts[19].

The advent of deep learning[20] significantly advanced segmentation performance. The U-Net[9] architecture, with its encoder-decoder structure and skip connections, became foundational for biomedical image segmentation. Extensions such as ResU-Net[13], which integrates residual blocks, and models with nested skip pathways and deep supervision, have further improved segmentation of complex retinal structures, including GA lesions.

Several studies have focused on applying deep learning to GA segmentation. A modified 3D U-Net applied to volumetric OCT data reported DSC between 0.66 and 0.89, with sensitivity ranging from 0.47 to 0.98 and specificity up to 0.99, highlighting its capacity to distinguish GA from surrounding tissue[21]. Another approach leveraging optical attenuation coefficients within a deep convolutional framework achieved a Dice score of $0.72 \pm 0.26$ and competitive sensitivity and specificity values[22]. A semi-automatic method combining region-growing, thresholding, and supervised refinement also achieved comparable performance while offering robustness to low-contrast images[23].

A comprehensive review reported that DSC for GA segmentation typically ranges between 0.68 and 0.89, with sensitivity between 0.47 and 0.98 and specificity from 0.73 to 0.99[24]. These variations are attributed to differences in imaging modality, dataset characteristics, and algorithmic design. More recently, deep learning models based on en-face swept-source OCT images have reported Dice scores approaching 0.89, with sensitivity and specificity comparable to expert graders[25]. U-Net–based methods on FAF images and en-face projections from volumetric SD-OCT data[26] have also demonstrated high segmentation accuracy and structural preservation[27].

In this study, we build upon these developments by comparing deep learning architectures with distinct modeling strategies for GA segmentation, ranging from classical threshold based methods to conventional CNN-based encoder-decoder networks to diffusion-based and transformer-based models by evaluating performance across these heterogeneous frameworks, we aim to identify robust solutions for automated GA segmentation and cross-sectional quantification, laying the groundwork for future progression analysis pipelines.

# 4. Methods

We conducted a systematic evaluation of several state-of-the-art models for GA segmentation. The following subsections describe the preprocessing pipeline, dataset composition, and model architecture in detail.

## 4.1 Dataset

The dataset used in this study comprises en-face OCT images acquired from the University of Pittsburgh Medical Center, encompassing a diverse and clinically representative collection of scans. The study was conducted in accordance with the Declaration of Helsinki and the Institutional Review Board (IRB: STUDY20030263) of the University of Pittsburgh. These images were obtained using two widely adopted imaging platforms: Cirrus® HD-OCT (Carl Zeiss Meditec, Inc., USA) and Spectralis® OCT (Heidelberg Engineering, Germany). The dataset comprises 269 en-face OCT images in total, with 146 acquired on Cirrus (54.3%) and 123 on Spectralis (45.7%), ensuring an approximately balanced device distribution across all splits. The dataset was divided into two parts patient-wise. The first part (~211 images, 80% of patients) was used for five-fold cross-validation, with each fold further divided into a training set (~168 images) and a fold-specific validation set (~43 images) used exclusively for early stopping and best checkpoint selection. The second part (58 images, 20% of patients) was held out as a fixed benchmark set, remaining entirely unseen throughout training and model selection, and constitutes the primary evaluation set reported in all results tables. Device balance was preserved across all splits, with approximately 55% Cirrus and 45% Spectralis images in each subset. Full split details and per-device distributions are provided in Tables 1 and 2. This heterogeneous distribution is critical for evaluating model robustness across varying imaging modalities, as the two devices differ in their acquisition protocols, resolution, and contrast characteristics. By incorporating data from both platforms, the dataset captures a broad range of real-world variability in image quality and anatomical presentation, ensuring that the models trained and validated on this data are more likely to generalize well to unseen clinical cases. The inclusion of both Spectralis scans and Cirrus scans makes this dataset particularly suitable for developing segmentation algorithms that are resilient to device-specific artifacts and contrast differences, thereby enhancing their clinical applicability.

**Table 1 Dataset modality split**

| Category | Image | Patient count |
|---|---|---|
| Cirrus | 146 | 76 |
| Spectralis | 123 | 99 |
| Total | 269 | 175 |

**Table 2 Dataset split**

| Split | Cirrus | Spectralis | Total |
|---|---|---|---|
| Training | 91 | 77 | 168 |
| Fold-specific validation | 23 | 20 | 43 |
| Held-out benchmark | 32 | 26 | 58 |

Ground truth segmentation masks were manually annotated by five ophthalmologists with varying levels of clinical experience, ranging from ophthalmology residents to consultant-level retinal surgeons, using a custom graphical annotation interface developed in-house. Each image was independently annotated by a single grader. Annotators were instructed to delineate the outer boundary of the GA lesion on the en-face OCT image, with reference to the corresponding B-scan stack where necessary to resolve ambiguous boundaries. All annotations underwent a data quality review to verify mask integrity, label consistency, and file completeness prior to inclusion in the dataset. Formal inter-grader agreement was not assessed, as each image received a single annotation, which represents a limitation of the current study.

## 4.2 Preprocessing

Preprocessing is a critical component of the proposed pipeline, as the quality of en-face images directly impacts segmentation performance. A key challenge in generating reliable en-face from volumetric OCT data is intra-volume contrast variability across B-scans. Differences in contrast and signal intensity introduce prominent horizontal line artifacts in reconstructed en-face images as shown in Fig 2b, which can obscure pathological features and degrade downstream segmentation accuracy.

**Fig 2b** En-face image generated by averaging pixel intensities from unnormalized B-scans.

To mitigate this issue, we employed a two-stage normalization strategy. First, a representative B-scan with optimal contrast was manually selected as a reference for all the volumes. All other B-scans were then histogram-matched to this reference image to align their intensity distributions. This normalization substantially reduced contrast inconsistencies across slices and suppressed line artifacts, resulting in a more uniform and visually consistent en-face image as shown in Fig 2c. To illustrate the sensitivity of en-face reconstruction to reference selection, en-face images generated using different reference B-scans are shown in Fig 3.

**Fig 2c** En-face image generated by averaging pixel intensities from histogram-matched B-scans.

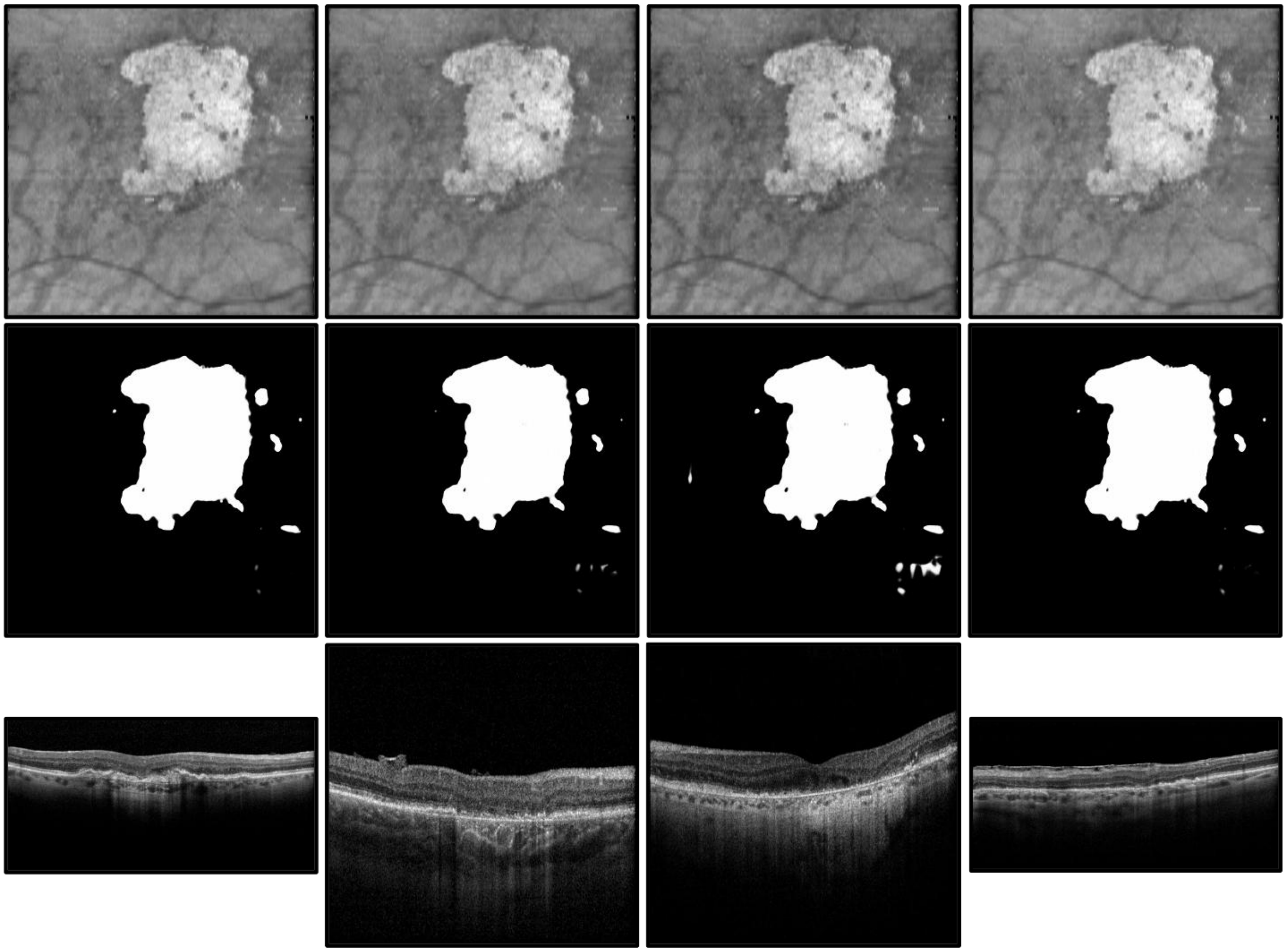

**Fig 3**. Histogram Matching with Various B-scans. (a) En-face image generated by histogram-matching all B-scans in a given volume to the reference B-scan shown in (c), followed by pixel-wise averaging. (b) Geographic atrophy (GA) segmentation overlaid on the en-face image from (a). (c) Reference B-scan used for histogram matching.

En-face projections were generated using a previously validated choroid segmentation algorithm[6]. For each row in the B-scan, the first pixel of the segmented choroidal region was identified, and a fixed depth of 400 μm below this boundary was selected as last point. This× depth was chosen because GA-related structural changes extend several hundred micrometers beneath the choroidal boundary in OCT images, making this range sufficient to consistently capture disease-associated alterations across subjects. Pixel intensities within this slab were averaged across all the rows to generate an en-face representation that provided a stable and anatomically consistent view of the region of interest.

In some cases, localized failures of choroid segmentation resulted in black line artifacts in the en-face images as illustrated in Fig 4a. Rather than discarding these scans, which would reduce dataset size and bias the evaluation, we applied a constrained inpainting strategy following prior work [28]. Inpainting was limited to small peripheral regions and to a maximum of two consecutive segmentation failures, ensuring that no large contiguous areas or lesion regions were altered. This targeted correction effectively removed visually disruptive artifacts while preserving underlying anatomical and pathological information, yielding cleaner en-face images suitable for model training and evaluation as shown in Fig 4b.

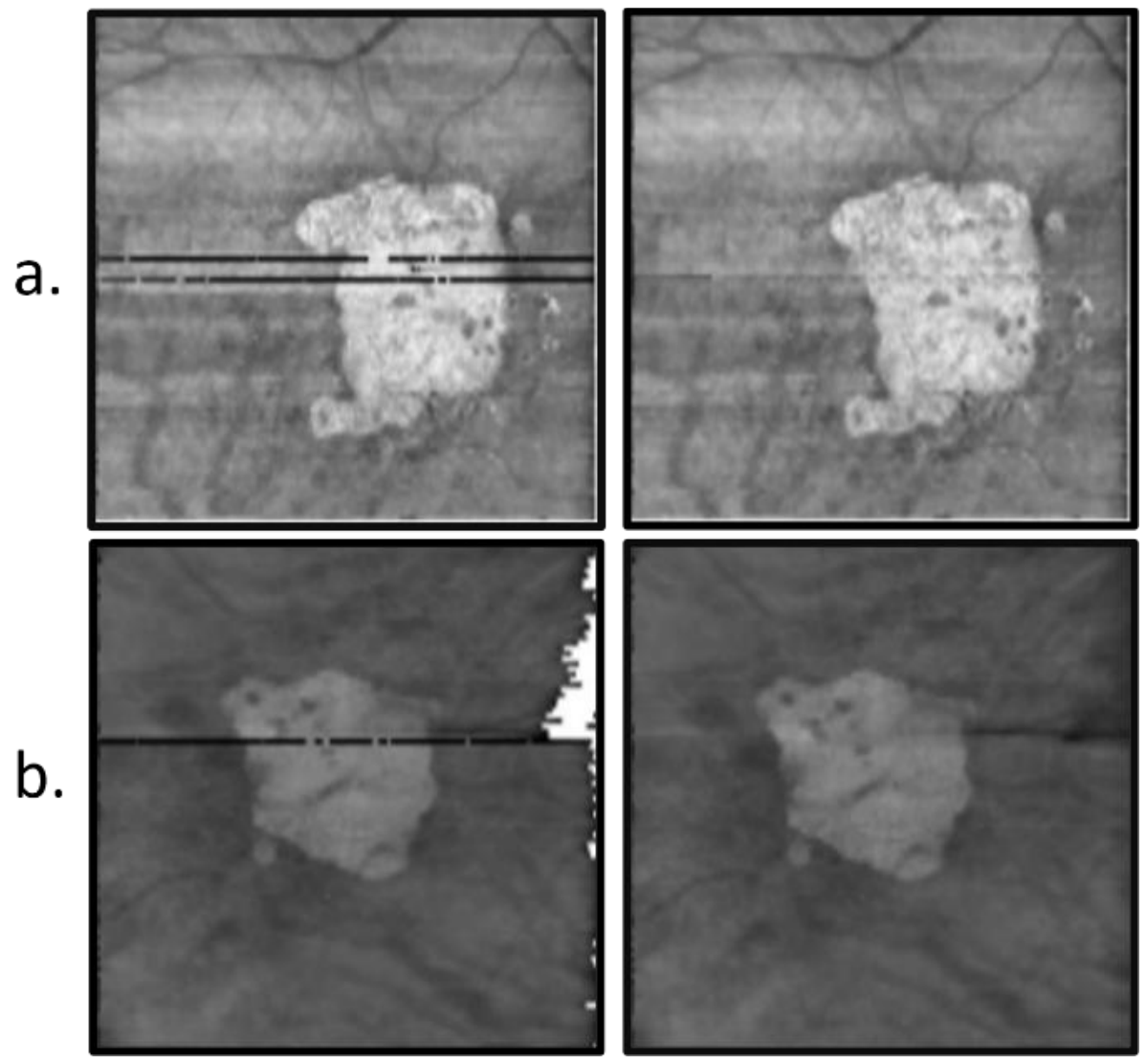

**Fig 4**. Inpainting. (a) Two en-face images exhibiting artifacts: black lines resulting from failed choroid segmentation and white noise introduced during B-scan extraction. (b) Inpainted en-face images with artifacts corrected.

Finally, all images were resized to 512×512 pixels prior to training, except for the Swin U-Net, which required a resolution of 448×448 due to architectural constraints and for MedSegDiff, which was trained on 256×256 owing to computational constraint. Standard data augmentation strategies like vertical flipping, contrast enhancement, and gamma correction were employed during training to enhance model robustness and simulate common variations encountered in clinical OCT data.

## 4.3 Models

We evaluated five segmentation models spanning CNNs, transformer-based and diffusion-based architectures. These are presented in Figs 5-9.

### 4.3.1 Nested U-Net

Nested U-Net as illustrated in Fig 5 extends the classical U-Net architecture by introducing dense nested skip connections across multiple resolution levels[14]. These connections facilitate multi-scale feature fusion and enhance the network's ability to recover fine-grained details, which is particularly important for accurately delineating the irregular and often ambiguous boundaries of GA lesions.

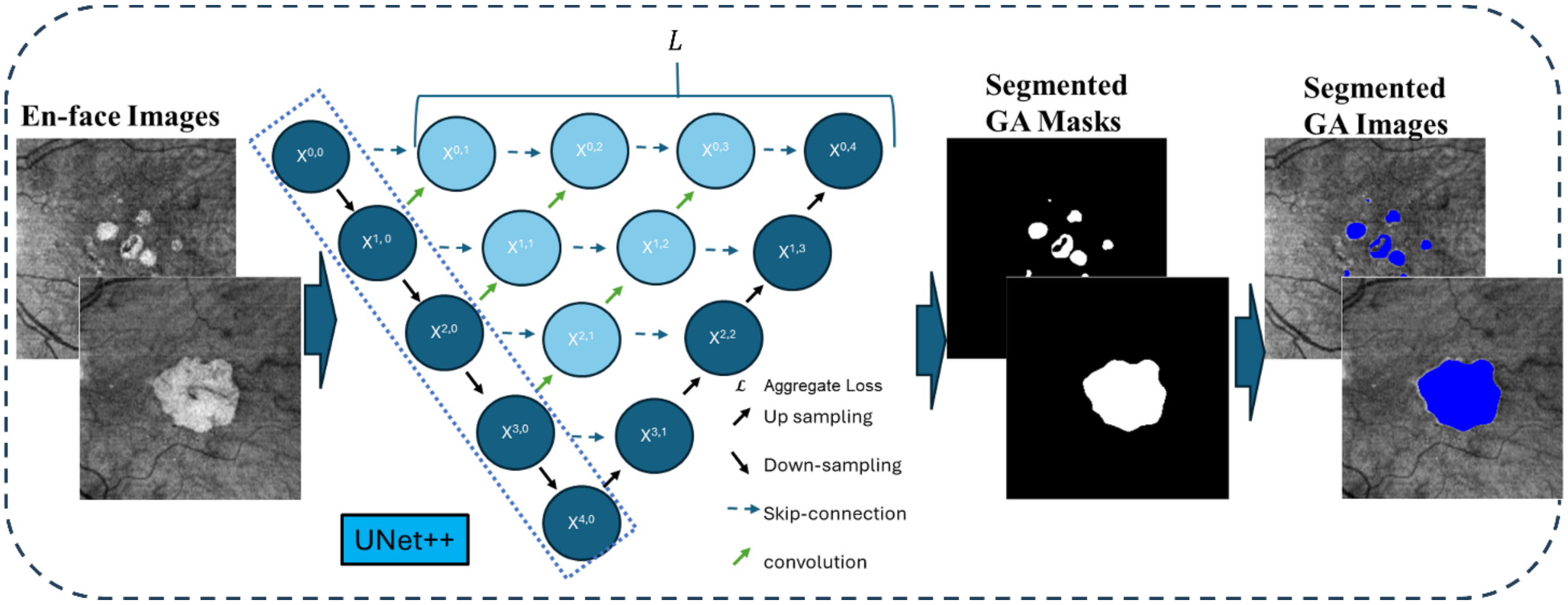

**Fig 5**. The Nested U-Net architecture is an extension of the standard U-Net model, incorporating nested and dense skip connections to better capture multi-scale contextual information and improve feature propagation.

To further improve learning, models incorporate deep supervision, a technique in which loss functions are applied to intermediate decoder layers in addition to the final output. By enforcing learning at multiple semantic levels, deep supervision promotes robust convergence and ensures that intermediate representations contribute meaningfully to the final segmentation.

### 4.3.2 ResU-Net

ResU-Net as illustrated in Fig 6 modifies the U-Net by incorporating residual blocks from the ResNet[29] architecture into both the encoder and decoder paths[13]. These residual connections alleviate vanishing gradient issues and improve feature propagation, allowing deeper networks to be trained efficiently. The use of residual units enhances the model's capacity to learn discriminative features in complex retinal images.

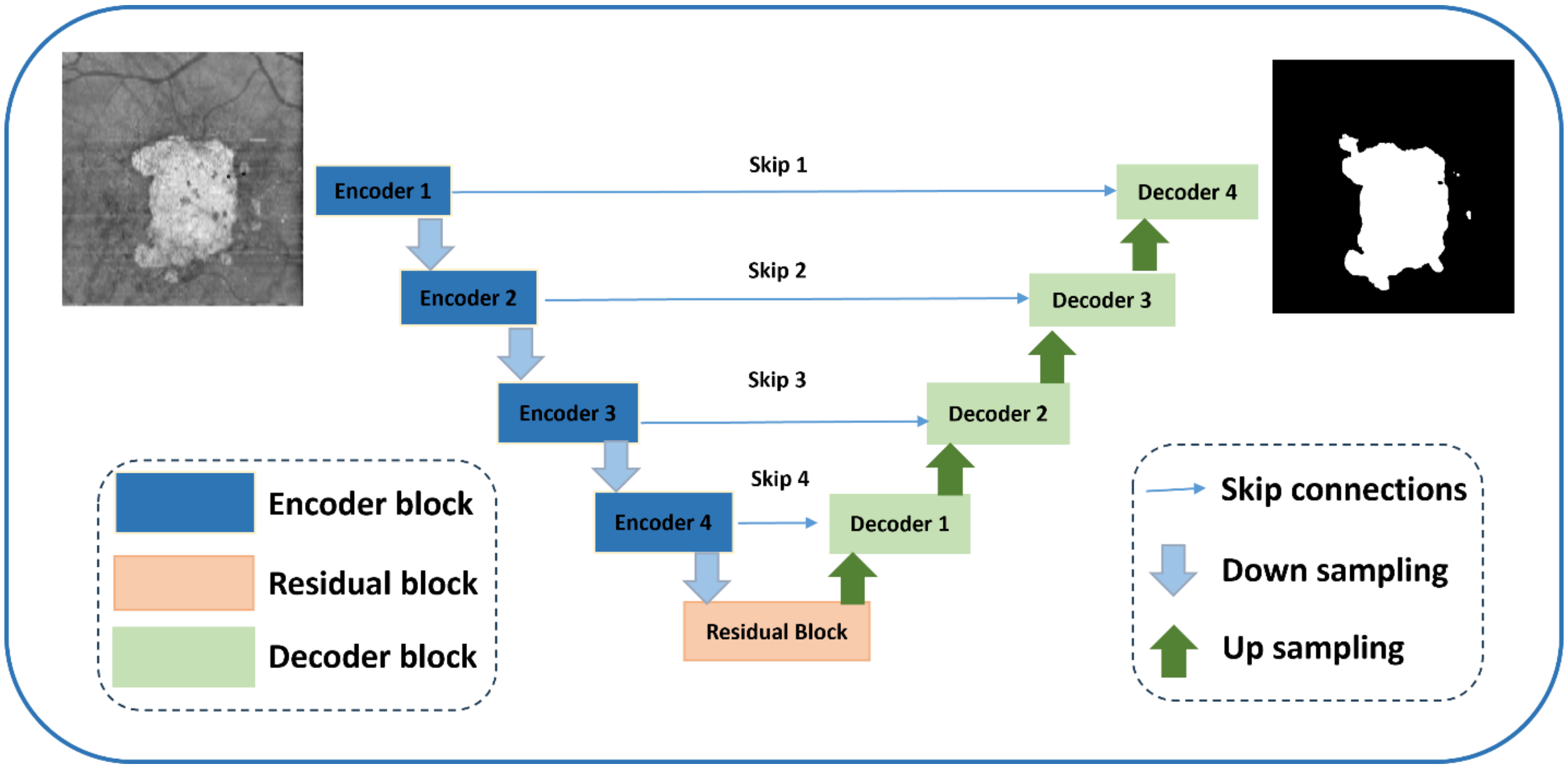

**Fig 6**. ResU-Net is an extension of the U-Net architecture that integrates residual blocks to facilitate gradient flow and improve feature learning in deep networks.

### 4.3.3 nnUnet

It is an automated segmentation framework that standardizes model design and training for medical image segmentation[12]. The framework itself doesn't have a fixed architecture. Instead, it derives key components such as network depth, width, preprocessing, data augmentation, and training schedules from the dataset. It automatically configures and trains the network and, because these steps are predefined and rule-based, it ensures reproducible segmentation pipelines without manual tuning. Architecture structure is illustrated in Fig 7.

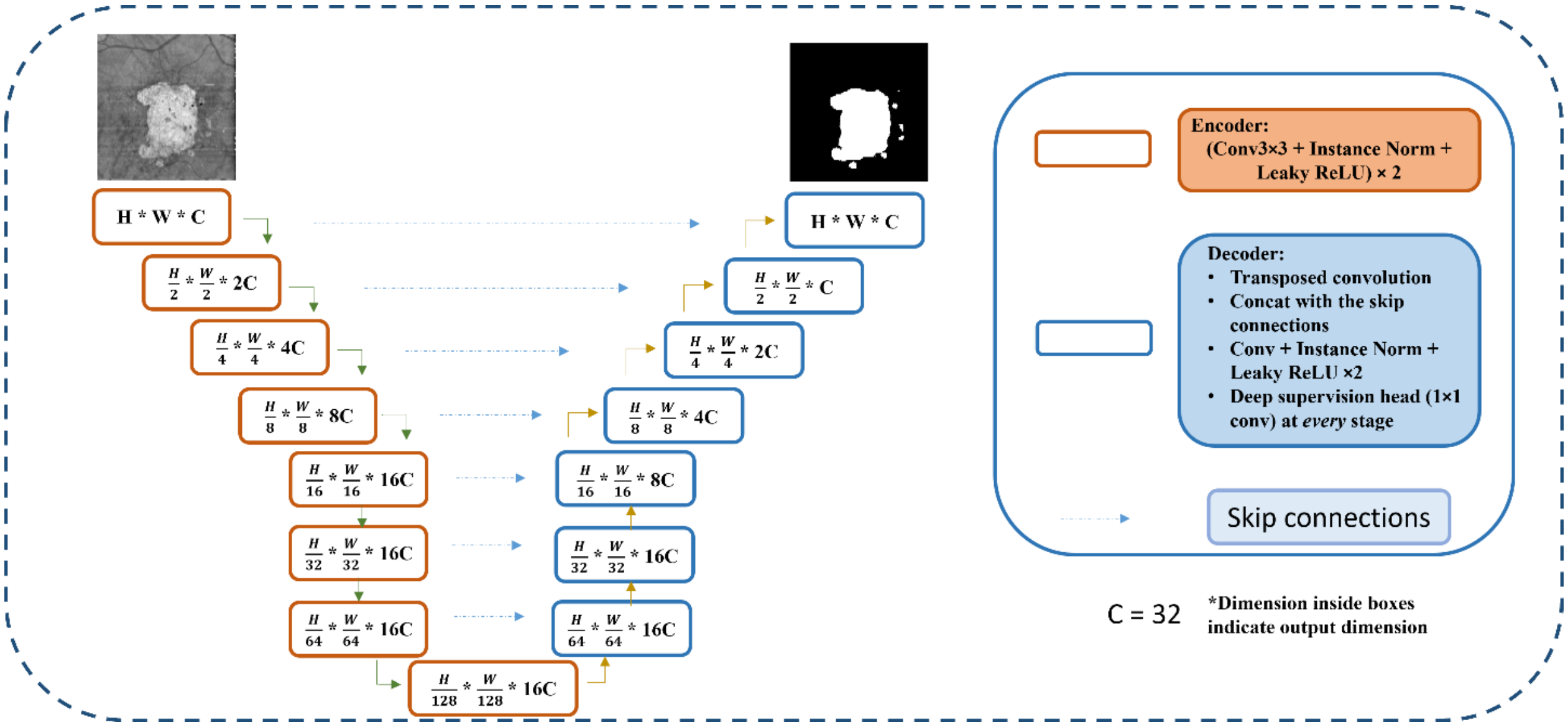

**Fig 7**. Architecture build using the nnUnet configurations. It consists of simple stacked convolutions.

### 4.3.4 Swin U-Net

Swin U-Net as illustrated in Fig 8 is a transformer-based segmentation model that replaces convolutional operations with Swin Transformer blocks[15]. These blocks use hierarchical feature representations and shifted window attention[30] to efficiently model both local and global spatial dependencies. With an encoder-decoder structure similar to U-Net, Swin U-Net combines spatial precision with contextual awareness, making it well-suited for dense prediction tasks in medical imaging.

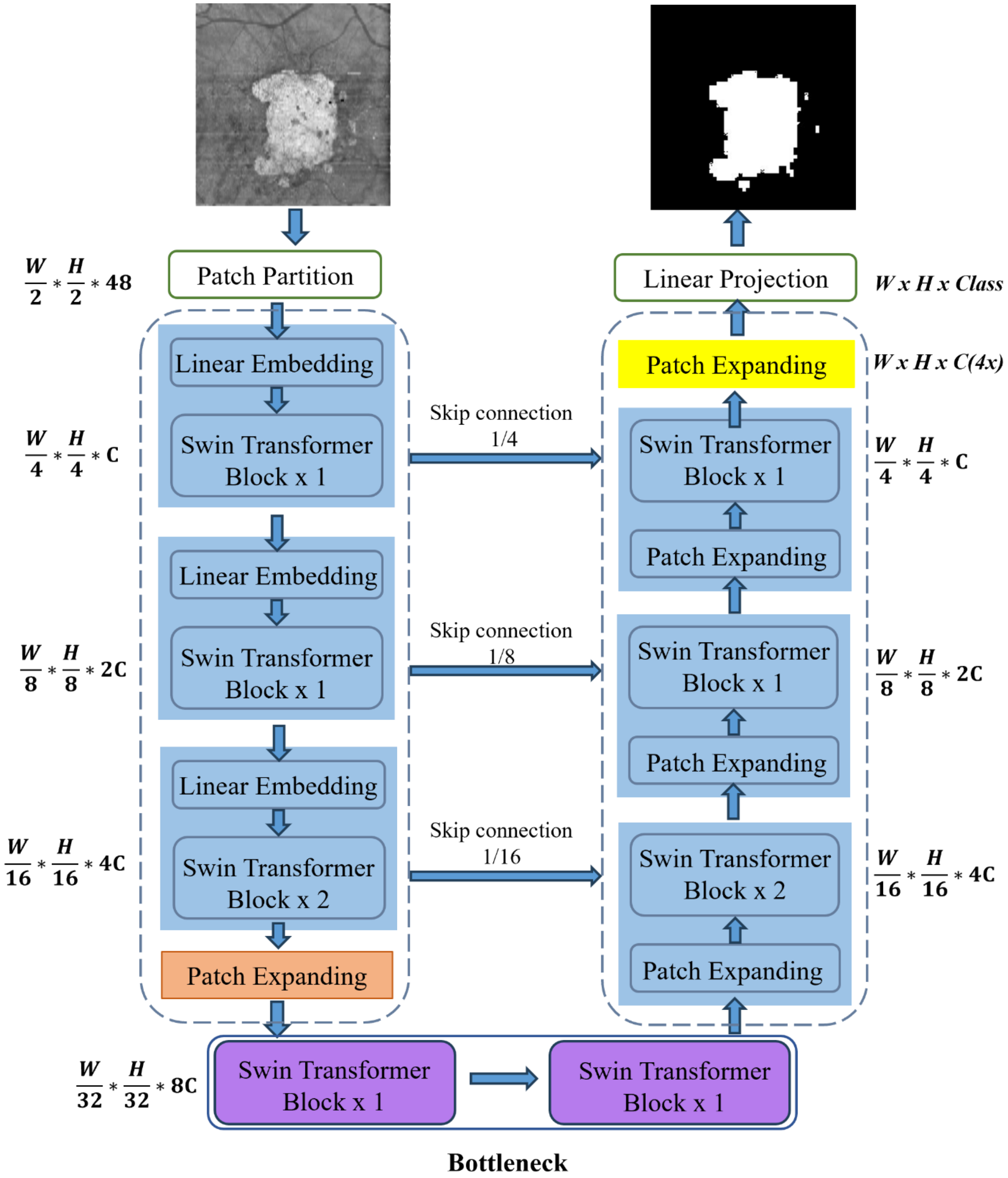

**Fig 8**. Swin U-Net is a transformer-based architecture that replaces conventional convolutional layers in the U-Net framework with hierarchical Swin Transformer blocks to better capture long-range dependencies and contextual information.

### 4.3.5 MedSegDiff Model

Diffusion models are a class of generative frameworks that learn to reverse a gradual noising process applied to data[31,32]. We used MedSegDiff[16] (v1) as illustrated in Fig 9, a diffusion-based image segmentation model that treats mask generation as a denoising task[16]. During training, noise is progressively added to ground truth masks, and the model learns to reconstruct the original masks conditioned on the input image. Diffusion models offer improved uncertainty estimation and sample diversity compared to deterministic models and have shown promise in high-fidelity medical image segmentation.

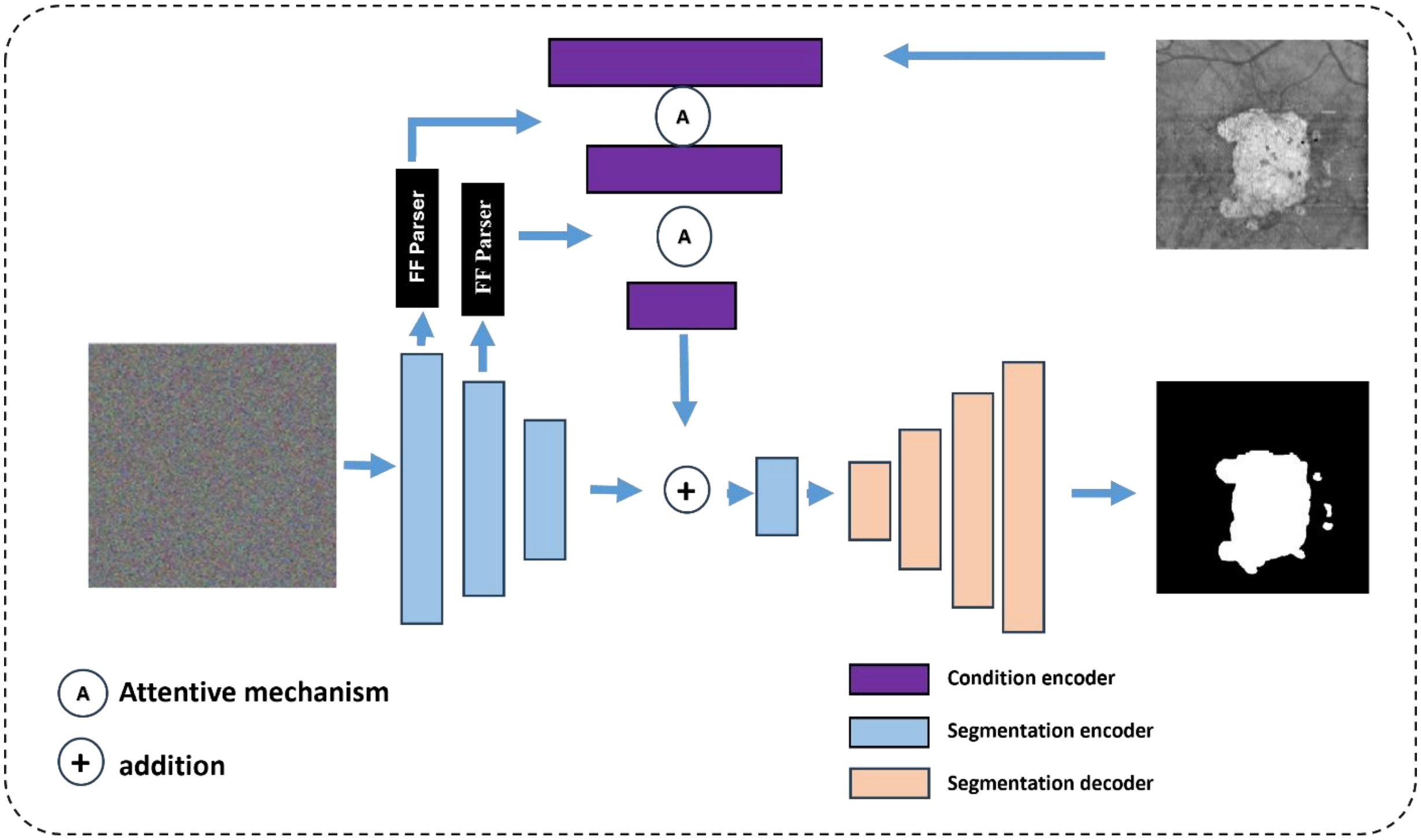

**Fig 9.** MedSegDiff is a diffusion-based model tailored for medical image segmentation. It incorporates an attention mechanism to dynamically localize and refine segmentation outputs, while the FF-Parser module suppresses high-frequency noise introduced during the integration of segmentation map features.

## 4.4 Classical Baselines

To quantify the contribution of the proposed preprocessing pipeline independently of deep learning, we evaluated five classical intensity-based segmentation methods on the preprocessed en-face OCT images: Otsu thresholding, an optimised global threshold, adaptive Gaussian thresholding, Niblack thresholding, and Sauvola thresholding. Parameter-dependent methods were optimised exclusively on the training partition of each fold and subsequently applied unchanged to the corresponding test partition, preventing information leakage.

## 4.5 Evaluation metric

To comprehensively evaluate segmentation performance, we report the Dice Similarity Coefficient (DSC), Intersection over Union, 95$^{th}$ percentile Hausdorff Distance, Sensitivity and Specificity.

**Dice Similarity Coefficient (DSC):**

It was used to quantify the agreement between automatically segmented geographically atrophy (GA) lesions and expert manual annotations. For each en-face OCT image, DSC measures the degree of spatial overlap between the binary GA mask predicted by the model and the corresponding ground-truth GA annotation delineated by trained graders. It is defined as

$$DSC = \frac{2|P \cap G|}{|P| + |G|}$$

Where P denotes the set of pixels classified as GA by the model and G denotes the set of pixels labeled as GA in the ground-truth annotation. The operator |.| represents the number of pixels in each set.

A DSC value of 1 indicates perfect agreement between the predicted and annotated GA regions, while a value of 0 indicates no overlap.

**Intersection over Union (IoU):**

IoU was used to evaluate the overlap between the automatically segmented GA lesions and the corresponding expert manual annotations. For each en-face OCT image, IOU quantifies the proportion of the overlapping region relative to the combined area of the predicted and ground-truth GA masks. It is defined as

Where P denotes the set of pixels classified as GA by the model, G denotes the set of pixels labeled as GA in the ground-truth annotations, and ∪ union of two sets.

Compared with Dice Similarity Coefficient, IoU provides a more stringent measure of segmentation accuracy by penalizing both false-positive and false-negative predictions through the union of the predicted and ground-truth regions.

**95th Percentile Hausdorff Distance (HD95):**

It was used to assess the boundary agreement between the automatically segmented GA lesions and the expert manual annotations. Unlike overlap-based metrics as DSC and IoU, HD95 measures the distance between the boundaries of the predicted and ground-truth segmentation masks. It is defined as the 95th percentile of the set of the shortest Euclidean distances between all boundary points of the predicted segmentation and the corresponding boundary points of the ground-truth segmentation.

$$HD_{95}(P,\, G) = \max\left(Q_{95}\left(\min_{g \in G} d(p,\, g)\right)_{p \in P},\ Q_{95}\left(\min_{p \in P} d(p,\, g)\right)_{g \in G}\right)$$

Where P and G denote the boundary points sets of the predicted and ground-truth segmentations, d(,. ,) represents the Euclidean distance between two boundary points, and $Q_{95}$ denotes the 95th percentile operator.

A lower value indicates closer agreement between the predicted and reference lesion boundaries, while a value of 0 corresponds to perfect boundary alignment. By excluding the most extreme 5% of boundary distances, HD95 is less sensitive to isolated outlier points than the conventional

Hausdorff Distance, making it a more robust and clinically meaningful metric for evaluating boundary accuracy in GA lesion segmentation.

**Sensitivity:**

Sensitivity, or the true positive rate, measures the proportion of GA pixels correctly identified by the model:

$$Sensitivity = \frac{TP}{TP + FN}$$

Where TP denotes correctly segmented GA pixels and FN denotes GA pixels missed by the model. A high sensitivity indicates that the model successfully captures the full spatial extent of GA lesions, minimizing missed regions.

**Specificity:**

Specificity, or the true negative rate, measures the proportion of healthy retinal pixels correctly classified as background:

$$Specificity = \frac{TN}{TN + FP}$$

Where TN denotes correctly identified background pixels and FP denotes healthy pixels incorrectly labelled as GA. A high specificity indicates that the model avoids over segmentation into surrounding healthy tissue, ensuring lesion area measurements are not artificially inflated.

# 5. Results

We evaluated the performance of five segmentation models- Nested U-Net, ResU-Net, nnUnet, Swin U-Net, and MedSegDiff and a few classical baselines-on a fixed held-out validation set comprising en-face OCT images from Cirrus and Spectralis devices. All models were implemented in PyTorch and trained on an NVIDIA RTX 2080 Ti GPU with 12 GB of VRAM.

ResU-Net and Nested U-Net were trained for 200 epochs using a cosine annealing learning rate schedule with an initial rate of $1x10^{-3}$ . The best model checkpoint was selected at the epoch achieving the highest Dice score on the fold-specific held-out test split. Model training was performed using Adam[33] optimizer. The loss function used during training was a weighted combination of binary cross-entropy (BCE) and Dice loss. This hybrid formulation optimizes both pixel-wise accuracy and overall region consistency, which is essential for mitigating class imbalance often encountered in medical image segmentation.

We used nnUnet's self-configuring 2D pipeline with its default architecture and preprocessing. The network was optimised using nnUnet's default combined Dice and cross-entropy loss and trained for 200 epochs following the framework's standard protocol. For each fold, the checkpoint with the best validation Dice observed during training was selected for evaluation. All other training settings - SGD optimiser with Nesterov momentum, an initial learning rate of 0.01 with polynomial decay, and 5-fold cross-validation with patient-level stratification matching the split used for all other models - followed nnUnet's defaults.

Swin-Unet, a transformer-based segmentation model, was trained from scratch to ensure a fair comparison under identical data conditions across all models. Input images were resized to 448x448 pixels which is the largest resolution under 512x512 compatible with the model's architectural constraint. Training was performed for 200 epochs using stochastic gradient descent (SGD) with momentum and a polynomial learning rate decay schedule, following the original repository's recommended practice. The loss function was weighted sum of cross-entropy (0.4) and Dice loss (0.6). For the MedSegDiff, training was performed for 2000 epochs. The model was optimized using Adam[33] optimizer with a fixed learning rate of 1e-4, ensuring consistent training and leveraging the strengths of the diffusion architecture in denoising and structure preservation.

In terms of computational efficiency, ResU-Net and Nested U-Net models trained in about two hour each, nnUnet and Swin-Unet took around three hours. MedSegDiff was significantly more resource-intensive, taking nearly 50 hours due to its iterative denoising process.

Among the evaluated classical methods, Otsu thresholding achieved the best overall performance (Dice 0.440 ± 0.000, HD95 2.358 ± 0.000 mm ). Its relatively high sensitivity (0.843) but low specificity (0.691) indicates that, although the preprocessing pipeline successfully enhances GA lesions as hyperintense regions, simple intensity thresholding substantially over-segments surrounding retinal tissue, resulting in poor boundary localization and overlap accuracy. The results for all five methods are shown in Table 3.

**Table 3 Classical Baseline on validation set**

| Method | Dice | IoU | HD95 (mm) | Sensitivity | Specificity |
|---|---|---|---|---|---|
| Otsu | **0.4400 ± 0.0000** | **0.3418 ± 0.0000** | **2.358 ± 0.000** | 0.8430 ± 0.0000 | **0.6910 ± 0.0000** |
| Optimal threshold | 0.3583 ± 0.0066 | 0.2448 ± 0.0063 | 2.614 ± 0.041 | 0.8114 ± 0.0344 | 0.4388 ± 0.0352 |
| Adaptive Gaussian | 0.2723 ± 0.0000 | 0.1712 ± 0.0000 | 2.463 ± 0.000 | 0.7482 ± 0.0000 | 0.4385 ± 0.0000 |
| Niblack | 0.3217 ± 0.0000 | 0.2075 ± 0.0007 | 2.485 ± 0.003 | 0.7032 ± 0.0187 | 0.6574 ± 0.0203 |
| Sauvola | 0.2958 ± 0.0000 | 0.1929 ± 0.0000 | 2.458 ± 0.000 | **0.8622 ± 0.0000** | 0.3779 ± 0.0000 |

Despite the effectiveness of the preprocessing pipeline in enhancing lesion contrast, all deep learning models consistently outperformed the classical baseline. The best-performing model, nnUnet, achieved a Dice score of 0.888, representing an improvement of 0.448 Dice points over Otsu thresholding, while reducing HD95 by approximately 76% ( 0.560 vs 2.358 mm). Even the lowest-performing deep learning model exceeded the classical baseline by a considerable margin. These results demonstrate that although preprocessing provides an effective foundation by improving lesion visibility, the majority of the segmentation performance is derived from learned representations that capture lesion morphology, spatial context, and imaging variability beyond simple intensity differences.

The quantitative results are summarized in Table 4-5.

**Table 4 (Fixed Validation set averaged across 5 folds)**

| **Model** | **Dice** | **IoU** | **HD95 (mm)** | **Sensitivity** | **Specificity** |
|---|---|---|---|---|---|
| Otsu (baseline) | 0.4400 ± 0.0000 | 0.3418 ± 0.0000 | 2.358 ± 0.000 | 0.8430 ± 0.0000 | 0.6910 ± 0.0000 |
| Nested U-Net | 0.7932 ± 0.0423 | 0.6935 ± 0.0511 | 1.077 ± 0.075 | 0.8368 ± 0.0498 | 0.9624 ± 0.0095 |
| ResU-Net | 0.8453 ± 0.0074 | 0.7634 ± 0.0086 | 0.699 ± 0.086 | **0.9112 ± 0.0077** | 0.9702 ± 0.0044 |
| Swin-UNet | 0.6890 ± 0.0294 | 0.5610 ± 0.0304 | 1.668 ± 0.108 | 0.7522 ± 0.0135 | 0.9523 ± 0.0151 |
| MedSegDiff | 0.7959 ± 0.0242 | 0.7210 ± 0.0236 | 0.853 ± 0.092 | 0.8834 ± 0.0218 | 0.9401 ± 0.0319 |
| nnU-Net | **0.8882 ± 0.0033** | **0.8205 ± 0.0034** | **0.560 ± 0.083** | 0.8775 ± 0.0069 | **0.9868 ± 0.0009** |

Values reported as mean ± standard deviation across five training folds, each evaluated on the same fixed held-out benchmark set.

**Table 5 (Test set results averaged across 5 folds)**

| **Model** | **Dice** | **IoU** | **HD95 (mm)** | **Sensitivity** | **Specificity** |
|---|---|---|---|---|---|
| Otsu (baseline) | 0.4133 ± 0.0615 | 0.3068 ± 0.0526 | 2.366 ± 0.184 | 0.8272 ± 0.0186 | 0.6881 ± 0.0308 |
| Nested U-Net | 0.8321 ± 0.0455 | 0.7360 ± 0.0547 | 0.958 ± 0.190 | 0.8333 ± 0.0305 | 0.9835 ± 0.0036 |
| ResU-Net | 0.8792 ± 0.0227 | 0.8013 ± 0.0294 | 0.674 ± 0.078 | **0.9238 ± 0.0194** | 0.9827 ± 0.0036 |
| Swin-UNet | 0.7045 ± 0.0234 | 0.5860 ± 0.0249 | 1.558 ± 0.135 | 0.7385 ± 0.0245 | 0.9620 ± 0.0123 |
| MedSegDiff | 0.8112 ± 0.0370 | 0.7360 ± 0.0425 | 0.731 ± 0.184 | 0.9054 ± 0.0334 | 0.9367 ± 0.0404 |
| nnU-Net | **0.8998 ± 0.0236** | **0.8329 ± 0.0324** | **0.433 ± 0.102** | 0.9041 ± 0.0254 | **0.9884 ± 0.0043** |

Mean ± standard deviation across five training folds, each evaluated on the fold-specific validation set. Across all evaluated metrics, nnUnet consistently achieved the strongest overall segmentation performance, attaining the highest Dice coefficient (0.888 ± 0.003), IoU (0.821 ± 0.003), specificity (0.987 ± 0.001) and lowest HD95 ( 0.560 ± 0.083 mm) on the fixed validation set, with remarkably low fold-to-fold variance indicating stable and reproducible behaviour across different patient subsets. ResU-Net ranked second in Dice (0.845 ± 0.007) while achieving the highest sensitivity (0.911 ± 0.008), suggesting a strong ability to detect the full spatial extent of GA lesions with minimal missed regions. Notably, Nested U-Net and Swin-Unet showed comparatively lower performance, with Swin-Unet achieving the lowest Dice (0.689 ± 0.029) and highest HD95 ( 1.668 ± 0.108 mm) among deep learning models, consistent with the known data-intensive requirements of transformer architectures when trained from scratch on small datasets. The Otsu threshold baseline, while achieving high sensitivity (0.843), exhibited low specificity (0.691) and a Dice of 0.440 - confirming that deep learning models provide a substantial and clinically meaningful improvement over intensity-based thresholding alone, with the best-performing model (nnUnet) achieving more than double the Dice score of the classical baseline.

Statistical comparison across models using the Friedman test revealed a significant overall difference in segmentation performance ($\chi^2$=19.04, $p<0.001$), confirming that the observed ranking reflects genuine performance differences rather than sampling variation. Device-stratified analysis further revealed that all models achieved consistently higher Dice scores on Spectralis than Cirrus images, with the significance and magnitude of this gap varying substantially across models.

To further examine cross-device generalisation, performance was analysed separately for Cirrus (n=32) and Spectralis (n=26) images in the validation set, with device differences assessed using the Mann-Whitney U test. nnUnet achieved the highest Dice on both Cirrus (0.876 ± 0.002) and Spectralis (0.903 ± 0.005), and the device gap did not reach significance ($\Delta$=+0.027, p=0.056), indicating robust cross-device generalisation. ResU-Net similarly showed no significant device dependence (Cirrus: 0.831 ± 0.008, Spectralis: 0.863 ± 0.008, $\Delta$=+0.033, p=0.321), making it the most device-robust among the non-self-configuring models. In contrast, Nested U-Net (Cirrus: 0.759 ± 0.058, Spectralis: 0.836 ± 0.023, Δ=+0.077, p=0.003) and MedSegDiff (Cirrus: 0.738 ± 0.041, Spectralis: 0.867 ± 0.013, Δ=+0.129, p=0.034) exhibited statistically significant performance gaps between devices, indicating sensitivity to the imaging source. Swin-Unet showed a moderate but non-significant gap (Δ=+0.043, p=0.467), consistent with its weaker overall performance. Notably, the models with the lowest fold-to-fold variance - ResU-Net (std=0.007) and nnUnet (std=0.003) - were also the most device-robust, suggesting that model stability and cross-device generalisation reflect the same underlying property: robustness to distribution shift. This finding has direct clinical relevance, as deployment across heterogeneous imaging equipment requires models that generalise reliably beyond the training distribution. Results for both the modalities are discussed in Table 6-9

**Table 6. Validation Set - Cirrus**

| **Model** | **Dice** | **IoU** | **HD95 (mm)** | **Sensitivity** | **Specificity** |
|---|---|---|---|---|---|
| Otsu (baseline) | 0.4330 ± 0.0000 | 0.3397 ± 0.0000 | 2.291 ± 0.000 | 0.8203 ± 0.0000 | 0.7058 ± 0.0000 |
| Nested U-Net | 0.7587 ± 0.0583 | 0.6397 ± 0.0704 | 1.289 ± 0.073 | 0.7869 ± 0.0849 | 0.9771 ± 0.0028 |
| ResU-Net | 0.8307 ± 0.0078 | 0.7358 ± 0.0095 | 0.731 ± 0.113 | **0.8944 ± 0.0109** | 0.9746 ± 0.0036 |
| Swin-UNet | 0.6696 ± 0.0355 | 0.5457 ± 0.0342 | 1.725 ± 0.244 | 0.7620 ± 0.0501 | 0.9422 ± 0.0284 |
| MedSegDiff | 0.7382 ± 0.0413 | 0.6508 ± 0.0383 | 1.080 ± 0.143 | 0.8544 ± 0.0234 | 0.9250 ± 0.0446 |
| nnU-Net | **0.8760 ± 0.0024** | **0.7941 ± 0.0027** | **0.588 ± 0.083** | 0.8462 ± 0.0093 | **0.9863 ± 0.0011** |

Validation set performance on Cirrus OCT images. Bold indicates best value per column. HD95 in mm .

**Table 7. Validation Set - Spectralis**

| **Model** | **Dice** | **IoU** | **HD95 (mm)** | **Sensitivity** | **Specificity** |
|---|---|---|---|---|---|
| Otsu (baseline) | 0.4485 ± 0.0000 | 0.3443 ± 0.0000 | 2.440 ± 0.000 | 0.8710 ± 0.0000 | 0.6728 ± 0.0000 |
| Nested U-Net | 0.8357 ± 0.0227 | 0.7598 ± 0.0275 | 0.817 ± 0.088 | 0.8981 ± 0.0143 | 0.9443 ± 0.0226 |
| ResU-Net | 0.8633 ± 0.0080 | 0.7973 ± 0.0083 | 0.659 ± 0.066 | 0.9318 ± 0.0042 | 0.9648 ± 0.0065 |
| Swin-UNet | 0.7129 ± 0.0273 | 0.5798 ± 0.0308 | 1.597 ± 0.125 | 0.7402 ± 0.0416 | 0.9649 ± 0.0036 |
| MedSegDiff | 0.8669 ± 0.0128 | 0.8074 ± 0.0153 | 0.574 ± 0.081 | **0.9191 ± 0.0210** | 0.9587 ± 0.0177 |
| nnU-Net | **0.9032 ± 0.0048** | **0.8531 ± 0.0050** | **0.525 ± 0.100** | 0.9161 ± 0.0059 | **0.9875 ± 0.0010** |

Validation set performance on Spectralis OCT images. Bold indicates best value per column. HD95 in mm .

**Table 8. Test Set - Cirrus**

| Model | Dice | IoU | HD95 (mm) | Sensitivity | Specificity |
|---|---|---|---|---|---|
| Otsu (baseline) | 0.5261 ± 0.0740 | 0.3958 ± 0.0665 | 2.086 ± 0.169 | 0.7654 ± 0.0203 | 0.7398 ± 0.0333 |
| Nested U-Net | 0.8240 ± 0.0380 | 0.7200 ± 0.0497 | 1.082 ± 0.272 | 0.7957 ± 0.0504 | 0.9838 ± 0.0047 |
| ResU-Net | 0.8840 ± 0.0300 | 0.8087 ± 0.0375 | 0.682 ± 0.148 | **0.9029 ± 0.0328** | 0.9785 ± 0.0067 |
| Swin-UNet | 0.7451 ± 0.0230 | 0.6276 ± 0.0234 | 1.444 ± 0.093 | 0.7810 ± 0.0418 | 0.9437 ± 0.0238 |
| MedSegDiff | 0.8389 ± 0.0566 | 0.7600 ± 0.0696 | 0.738 ± 0.190 | 0.8959 ± 0.0251 | 0.9314 ± 0.0521 |
| nnU-Net | **0.9005 ± 0.0342** | **0.8340 ± 0.0508** | **0.407 ± 0.109** | 0.8789 ± 0.0452 | **0.9865 ± 0.0043** |

Test set performance on Cirrus OCT images. Bold indicates best value per column. HD95 in mm .

**Table 9. Test Set - Spectralis**

| Model | Dice | IoU | HD95 (mm) | Sensitivity | Specificity |
|---|---|---|---|---|---|
| Otsu (baseline) | 0.2801 ± 0.0567 | 0.2016 ± 0.0458 | 2.696 ± 0.197 | 0.8999 ± 0.0190 | 0.6268 ± 0.0359 |
| Nested U-Net | 0.8422 ± 0.0601 | 0.7552 ± 0.0670 | 0.810 ± 0.245 | 0.8765 ± 0.0341 | 0.9834 ± 0.0102 |
| ResU-Net | 0.8741 ± 0.0362 | 0.7934 ± 0.0464 | 0.661 ± 0.165 | **0.9487 ± 0.0149** | 0.9878 ± 0.0024 |
| Swin-UNet | 0.6570 ± 0.0430 | 0.5370 ± 0.0463 | 1.692 ± 0.315 | 0.6879 ± 0.0436 | 0.9837 ± 0.0056 |
| MedSegDiff | 0.7783 ± 0.0230 | 0.7075 ± 0.0227 | 0.725 ± 0.202 | 0.9170 ± 0.0535 | 0.9430 ± 0.0303 |
| nnU-Net | **0.8990 ± 0.0363** | **0.8316 ± 0.0451** | **0.462 ± 0.117** | 0.9337 ± 0.0118 | **0.9906 ± 0.0046** |

Test set performance on Spectralis OCT images. Bold indicates best value per column. HD95 in mm .

**Table 10 Results from other approaches**

| Approach** | Dice Score |
|---|---|
| Chu, Z *et al.*[22] | 0.94 ± 0.03 |
| Anegondi, N. *et al.*[26] | 0.89-0.92 * |
| Liefers, B *et al.*[34] | 0.72 ± 0.26 |
| Niu, S *et al.*[18] | 0.87 ± 0.06 |

Results for different approaches proposed by other authors are given in Table 10. These externally reported results reflect the experimental setups of their respective studies and are included for reference rather than direct comparison.

Qualitative comparison of the model's performance is presented in Fig 10 and 11, where the segmentation masks generated by each model are overlaid on the corresponding en-face OCT images. This visual comparison further corroborates the quantitative results, with nnUnet producing the most accurate and stable segmentations across different device sources.

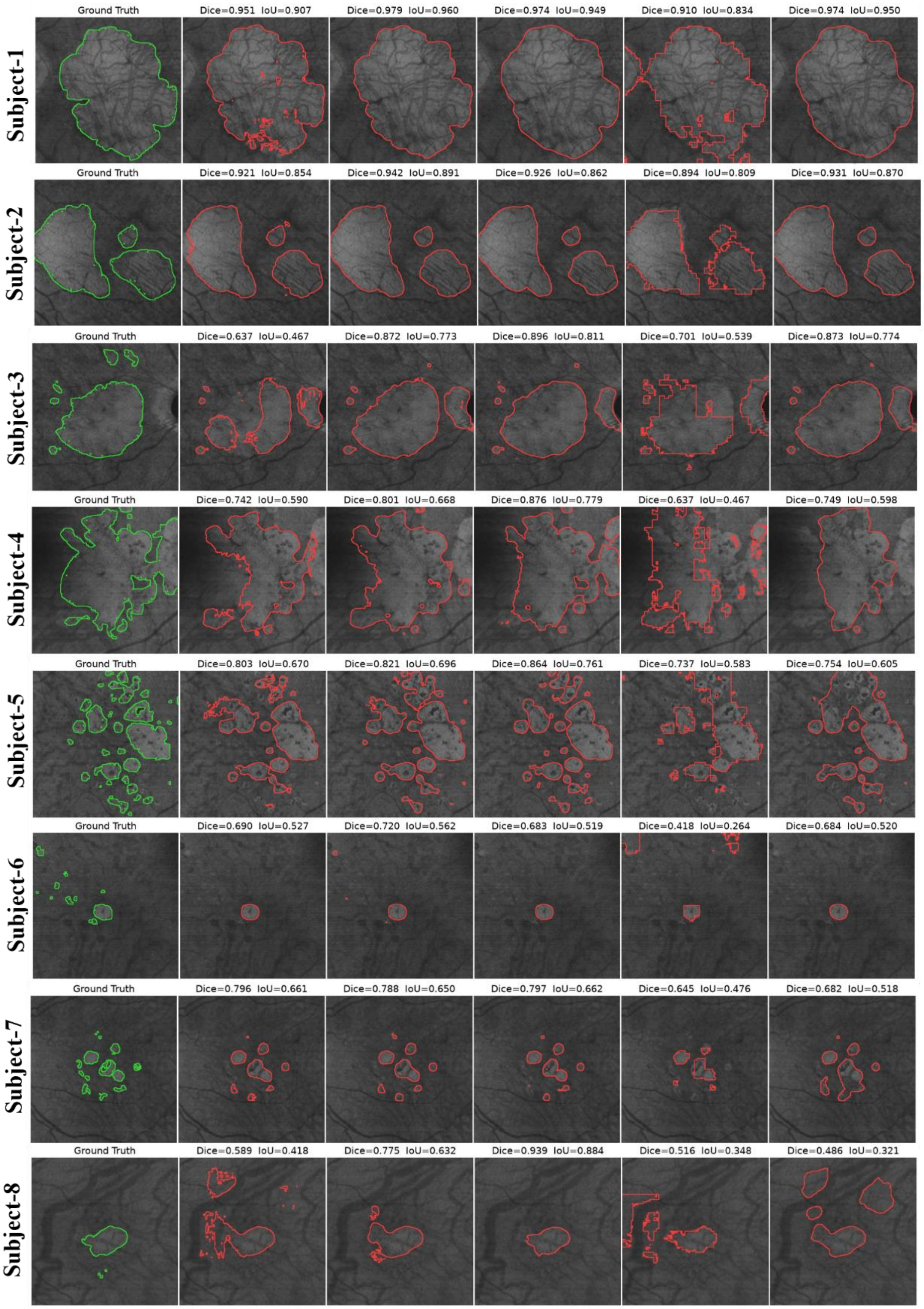

**Fig 10**. Segmentation results overlaid on cirrus en-face images for all five models across eight different subjects.

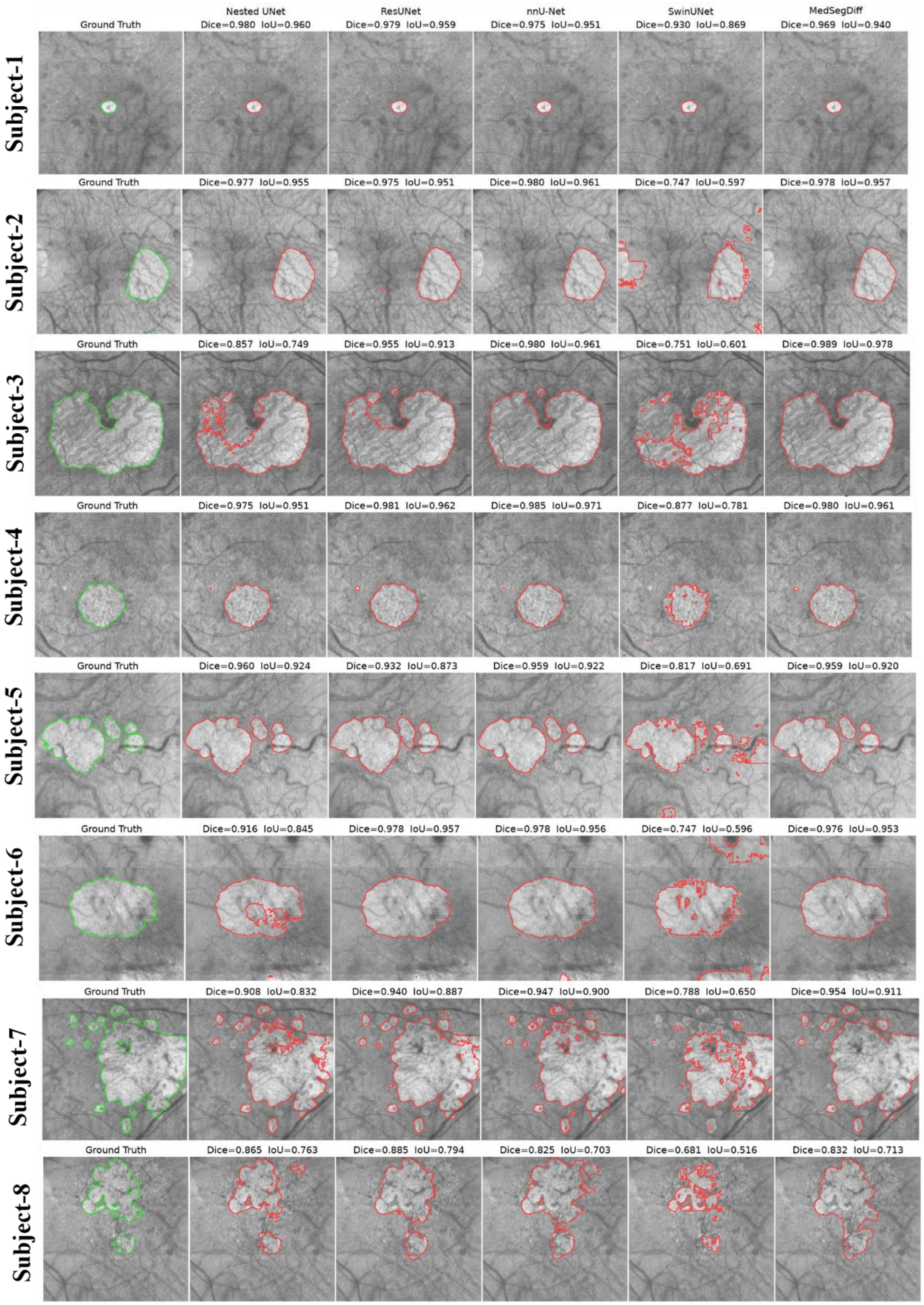

**Fig 11**. Segmentation results overlaid on spectralis en-face images for all five models across eight different subjects.

GA lesion area agreement between predicted and ground truth segmentation masks was assessed using mean absolute error (MAE) and Bland-Altman analysis, with all masks normalised to a canonical 512×512 resolution prior to computation. Since all en-face images represent a 6×6 mm field of view and all masks were resized to a canonical 512×512 resolution prior to area computation, a conversion factor of $(6/512)^2 \approx 1.373 \times 10^{-4}$ mm²/pixel² was applied uniformly across both devices. nnU-Net demonstrated the strongest area agreement, with an MAE of 0.487 mm² (3,541 pixels²) and a near-zero bias of −0.091 mm² (−664 pixels²), with Bland-Altman limits of agreement of [−1.971, +1.789] mm² ([−14,345, +13,018] pixels²), indicating minimal systematic over- or under-estimation of lesion area. ResU-Net showed a consistent positive bias of +0.494 mm² (+3,597 pixels²), suggesting a tendency to slightly overestimate lesion area. Nested U-Net and SwinUNet exhibited wider limits of agreement, reflecting greater variability in area estimation. Bland-Altman plots for all models are provided in Figure 12.

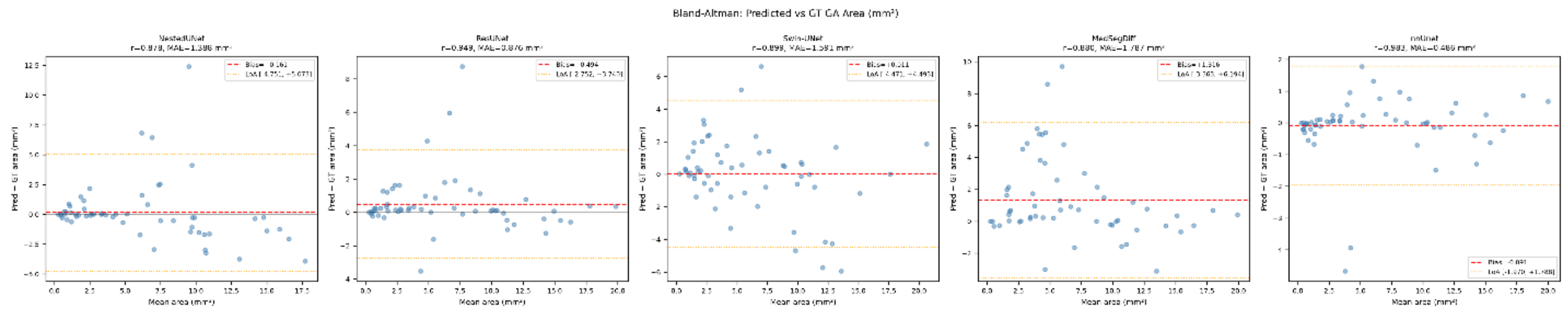

**Fig 12**. Bland-Altman analysis of GA lesion area agreement between predicted and ground truth segmentation masks across all five models. For each image, predicted and ground truth areas (pixels²) were averaged across five folds. The red dashed line indicates the mean bias and the orange dotted lines indicate the 95% limits of agreement (bias ± 1.96 SD). All areas computed at canonical 512×512 resolution.

These limits of agreement (~3.76 mm² total span for nnU-Net) exceed the pooled annual natural-history GA growth rate of 1.66 mm²/year reported by Shen et al. and are substantially larger than the 12-month absolute treatment effects of 0.21–0.41 mm² observed in the OAKS and DERBY trials. Accordingly, while the limits of agreement preclude fully automated individual-level progression monitoring, the pipeline is well-suited for cross-sectional lesion quantification and as an assistive tool in longitudinal studies, where clinician review and correction of predicted masks preserves measurement accuracy while substantially reducing manual annotation burden.

# 6. Discussion

Our comparative analysis demonstrates that, under a limited-data regime, nnUnet provides the most reliable performance for GA segmentation on en-face OCT, achieving the highest Dice score (0.888 ± 0.003) with remarkably low fold-to-fold variance, indicating stable and reproducible segmentation across patient subsets. ResU-Net ranked second in Dice (0.845 ± 0.007) while achieving the highest sensitivity (0.911 ± 0.008), benefiting from residual connections that stabilise gradient flow and support consistent convergence. MedSegDiff and Nested U-Net performed comparably in Dice (0.796 ± 0.024 and 0.793 ± 0.042 respectively). These results underscore the effectiveness of carefully designed architectures in medical image segmentation, particularly when training data is limited.

Unlike the other four models, which were trained using fixed hyperparameters following their authors' recommended protocols, nnUnet automatically configures its preprocessing, augmentation, architecture, and optimisation schedule directly from the dataset. This self-configuration represents an inherent asymmetry - the remaining models were not subjected to systematic hyperparameter optimisation, and their performance may improve with dedicated tuning. The reported results therefore reflect out-of-the-box performance under practical deployment conditions rather than the theoretical ceiling of each architecture.

Transformer-based models like Swin U-Net capture long-range dependencies and global morphology through self-attention mechanisms but also introduce subtle blocking artifacts due to the shifted-window design and require input resizing that degrades resolution.

The MedSegDiff's long training time (~50 hours) highlights the current limitations in computational efficiency. Improvements in inference-time sampling and training optimization could help make diffusion models more practical in medical imaging workflows.

On the preprocessing front, our three-stage histogram-matching preprocessing method reliably normalized inter-scan contrast and suppressed device-specific artifacts, supporting consistent training across architectures. Still, domain shifts between imaging platforms and limited resolution standardization remain challenging.

Beyond cross-sectional segmentation, longitudinal quantification of GA progression has become a critical focus in both clinical research and therapeutic trials. Techniques for tracking GA growth over time have been proposed[34], providing frameworks to measure atrophic area expansion. Automated pipelines for longitudinal GA quantification have been developed and validated on large cohorts[35] , and recent models[36] incorporate deep learning to improve temporal consistency and measurement precision.

Despite strong overall performance, the proposed pipeline exhibits several systematic failure modes. First, GA segmentation performance is highly dependent on the accuracy of the preceding choroid segmentation used for en-face generation. Errors or discontinuities in choroid segmentation can lead to incorrect slab selection, distorted en-face projections, and consequently degraded GA segmentation. This effect is particularly pronounced in cases with large or advanced GA lesions extending beyond the effective axial coverage of the OCT volume, where en-face representations fail to fully capture lesion extent.

Additionally, eye motion, and acquisition-related artifacts introduce misalignment across adjacent B-scans, resulting in spatial inconsistencies and boundary discontinuities in the generated en-face images. Such artifacts propagate through the pipeline and can produce noisy or incomplete segmentation labels, limiting model robustness. Other image quality issues, including shadowing, signal drop-off, and device-specific artifacts, further contribute to segmentation errors in challenging cases.

This study has several other important limitations. The dataset size was modest, particularly for training data-hungry models like transformers and diffusion networks. Second, the segmentation masks were manually annotated by five ophthalmologists with varying levels of clinical experience over the course of dataset development. Although each image was independently annotated using a custom annotation interface, inter-grader agreement was not formally assessed, which may have introduced variability in the reference annotations. Third, only en-face OCT images were used, limiting the ability to exploit rich volumetric or multimodal information. Fourth, MedSegDiff and Swin-Unet predictions were generated at a lower native resolution due to computational constraints, and while all masks were normalised to a 512×512 before metric computation, resolution-dependent segmentation quality differences may persist. Finally, the comparison carries an inherent asymmetry in that nnU-Net automatically configures its preprocessing, augmentation, and optimisation schedule from the data, while all other models were trained with fixed hyperparameters following their authors' recommended protocols. Systematic hyperparameter optimisation of the non-self-configuring models could potentially narrow the observed performance gap.

# 7. Conclusion

We have presented a head-to-head evaluation of five deep learning architectures for geographic atrophy (GA) segmentation in en-face OCT, encompassing convolutions, nested skip connection models, residual encoder–decoder networks, transformer-based architectures, and diffusion-based models. Our study demonstrates that, under a limited-data regime, CNNs with appropriate regularization offer the best trade-off between segmentation accuracy and training stability. nnUnet achieved the highest mean Dice score (0.888 ± 0.003), while ResU-Net, and Nested U-Net also showed competitive results with slight increases in variance or reduced resolution fidelity.

MedSegDiff provided high-fidelity boundary delineation and uncertainty estimates but suffered from greater performance variability and significantly higher computational costs. While promising in capturing lesion irregularities, it required substantially longer training time (~50 hours), limiting its practicality in time-sensitive or resource-constrained environments.

A key strength of our comparison lies in the robust three-stage preprocessing pipeline-comprising of B-scan histogram matching, choroid-based en-face construction and inpainting, which proved critical in normalizing device-specific contrasts and mitigating structural artifacts. Our findings also underscore the value of architectural choices: residual skip connections enhance gradient flow and convergence, while adversarial and transformer-based methods may require more data and stronger regularization to achieve consistent outcomes.

Overall, this work contributes a systematic benchmark for GA segmentation in en-face OCT images, offering the following key takeaways:

- Regularized CNNs can outperform complex architectures.
- Preprocessing strategies such as inpainting and histogram normalization play a pivotal role in improving image quality.

- Future work should explore ensemble methods, integration of volumetric and multi-modal imaging data, and self-supervised learning to further enhance performance in low-data settings. Our study lays the groundwork for the development of fast, and robust GA segmentation tool to support earlier diagnosis and monitoring in clinical practice.

## References

1 Bakri, S. J. *et al.* Geographic atrophy: Mechanism of disease, pathophysiology, and role of the complement system. *Journal of Managed Care & Specialty Pharmacy* **29**, S2-S11 (2023). https://doi.org/10.18553/jmcp.2023.29.5-a.s2

2 Fleckenstein, M., Schmitz-Valckenberg, S. & Chakravarthy, U. Age-Related Macular Degeneration: A Review. *JAMA* **331**, 147-157 (2024). https://doi.org/10.1001/jama.2023.26074

3 Bressler, N. M., Bressler, S. B. & Fine, S. L. Age-related macular degeneration. *Survey of Ophthalmology* **32**, 375-413 (1988). https://doi.org/10.1016/0039-6257(88)90052-5

4 Chatzara, A. *et al.* Artificial Intelligence Approaches for Geographic Atrophy Segmentation: A Systematic Review and Meta-Analysis. *Bioengineering* **12**, 475 (2025).

5 Sayegh, R. G. *et al.* A systematic comparison of spectral-domain optical coherence tomography and fundus autofluorescence in patients with geographic atrophy. *Ophthalmology* **118**, 1844-1851 (2011). https://doi.org/10.1016/j.ophtha.2011.01.043

6 Arora, S. *et al.* Three-Dimensional Choroidal Contour Mapping in Central Serous Chorioretinopathy and Fellow Eyes. *Translational Vision Science & Technology* **14**, 23-23 (2025). https://doi.org/10.1167/tvst.14.11.23

7 LeCun, Y., Bottou, L., Bengio, Y. & Haffner, P. in *Proceedings of the IEEE* Vol. 86 2278--2324 (1998).

8 Enzendorfer, M. L. & Schmidt-Erfurth, U. Artificial intelligence for geographic atrophy: pearls and pitfalls. *Current Opinion in Ophthalmology* **35**, 455-462 (2024). https://doi.org/10.1097/icu.0000000000001085

9 Ronneberger, O., Fischer, P. & Brox, T. in *Medical image computing and computer-assisted intervention–MICCAI 2015: 18th international conference, Munich, Germany, October 5-9, 2015, proceedings, part III 18.* 234-241 (Springer).

10 Yao, W. *et al.* From CNN to Transformer: A Review of Medical Image Segmentation Models. *Journal of Imaging Informatics in Medicine* **37**, 1529 - 1547 (2023).

11 Kazerouni, A. *et al.* Diffusion models in medical imaging: A comprehensive survey. *Medical image analysis* **88**, 102846 (2022).

12 Isensee, F. *et al.* in *International Conference on Medical Image Computing and Computer-Assisted Intervention.* 488-498 (Springer).

13 Diakogiannis, F. I., Waldner, F., Caccetta, P. & Wu, C. ResUNet-a: A deep learning framework for semantic segmentation of remotely sensed data. *ISPRS Journal of Photogrammetry and Remote Sensing* **162**, 94-114 (2020).

14 Zhou, Z., Rahman Siddiquee, M. M., Tajbakhsh, N. & Liang, J. 3-11 (Springer).

15 Cao, H. *et al.* 205-218 (Springer).
16 Wu, J. *et al.* in *Medical Imaging with Deep Learning.* 1623-1639 (PMLR).
17 Zhihong, H., Gerard, G. M., Matthias, H. & Srinivas, R. S. Automated segmentation of geographic atrophy in fundus autofluorescence images using supervised pixel classification. *Journal of Medical Imaging* **2**, 014501 (2015). https://doi.org/10.1117/1.JMI.2.1.014501
18 Niu, S., de Sisternes, L., Chen, Q., Leng, T. & Rubin, D. L. Automated geographic atrophy segmentation for SD-OCT images using region-based C-V model via local similarity factor. *Biomed. Opt. Express* **7**, 581-600 (2016). https://doi.org/10.1364/BOE.7.000581
19 Schütze, C. *et al.* Performance of OCT segmentation procedures to assess morphology and extension in geographic atrophy. *Acta Ophthalmologica* **89**, 235-240 (2011). https://doi.org/https://doi.org/10.1111/j.1755-3768.2010.01955.x
20 LeCun, Y., Bengio, Y. & Hinton, G. Deep learning. *Nature* **521**, 436-444 (2015). https://doi.org/10.1038/nature14539
21 Yoshida, K. *et al.* Deep Learning Approaches to Predict Geographic Atrophy Progression Using Three-Dimensional OCT Imaging. *Translational Vision Science & Technology* **14**, 11-11 (2025). https://doi.org/10.1167/tvst.14.2.11
22 Chu, Z. *et al.* Automatic geographic atrophy segmentation using optical attenuation in OCT scans with deep learning. *Biomed. Opt. Express* **13**, 1328-1343 (2022). https://doi.org/10.1364/BOE.449314
23 Chen, Q. *et al.* Semi-automatic geographic atrophy segmentation for SD-OCT images. *Biomed. Opt. Express* **4**, 2729-2750 (2013). https://doi.org/10.1364/BOE.4.002729
24 Arslan, J. *et al.* Artificial Intelligence Algorithms for Analysis of Geographic Atrophy: A Review and Evaluation. *Translational Vision Science & Technology* **9**, 57-57 (2020). https://doi.org/10.1167/tvst.9.2.57
25 Pramil, V. *et al.* A Deep Learning Model for Automated Segmentation of Geographic Atrophy Imaged Using Swept-Source OCT. *Ophthalmology Retina* **7**, 127-141 (2023). https://doi.org/10.1016/j.oret.2022.08.007
26 Anegondi, N. *et al.* Deep Learning to Predict Geographic Atrophy Area and Growth Rate from Multimodal Imaging. *Ophthalmology Retina* **7**, 243-252 (2023). https://doi.org/10.1016/j.oret.2022.08.018
27 Spaide, T. *et al.* Geographic Atrophy Segmentation Using Multimodal Deep Learning. *Translational Vision Science & Technology* **12**, 10-10 (2023). https://doi.org/10.1167/tvst.12.7.10
28 Suvorov, R. *et al.* 2149-2159.
29 He, K., Zhang, X., Ren, S. & Sun, J. in *Proceedings of 2016 IEEE Conference on Computer Vision and Pattern Recognition* 770--778 (IEEE, 2016).
30 Liu, Z. *et al.* Swin Transformer: Hierarchical Vision Transformer using Shifted Windows. *2021 IEEE/CVF International Conference on Computer Vision (ICCV)*, 9992-10002 (2021).
31 Dhariwal, P. & Nichol, A. in *Proceedings of the 35th International Conference on Neural Information Processing Systems* Article 672 (Curran Associates Inc., 2021).
32 Nichol, A. Q. & Dhariwal, P. in *ICML* Vol. 139 (eds Marina Meila & Tong Zhang) 8162-8171 (PMLR, 2021).
33 Kingma, D. P. & Ba, J. Adam: A method for stochastic optimization. *arXiv preprint arXiv:1412.6980* (2014).
34 Liefers, B. *et al.* A Deep Learning Model for Segmentation of Geographic Atrophy to Study Its Long-Term Natural History. *Ophthalmology* **127**, 1086-1096 (2020). https://doi.org/10.1016/j.ophtha.2020.02.009

35 Pilotto, E. *et al.* En Face Optical Coherence Tomography to Detect and Measure Geographic Atrophy. *Investigative Ophthalmology & Visual Science* **56**, 8120-8124 (2015). https://doi.org/10.1167/iovs.15-17366
36 Mai, J. *et al.* Clinical validation for automated geographic atrophy monitoring on OCT under complement inhibitory treatment. *Scientific Reports* **13**, 7028 (2023).